\documentclass[letterpaper]{article} 
\usepackage{aaai23}
\usepackage{times}  
\usepackage{helvet}  
\usepackage{courier}  
\usepackage[hyphens]{url}  
\usepackage{graphicx} 
\usepackage{natbib}  
\usepackage{caption} 
\usepackage{algorithm}
\usepackage{algorithmic}
\usepackage{defs}
\usepackage{lipsum}
\usepackage{xcolor}
\usepackage{float}
\usepackage{booktabs}
\usepackage{multirow}
\usepackage[normalem]{ulem}
\useunder{\uline}{\ul}{}
\usepackage{newfloat}
\usepackage{listings}
\DeclareCaptionStyle{ruled}{labelfont=normalfont,labelsep=colon,strut=off} 
\floatstyle{ruled}
\newfloat{listing}{tb}{lst}{}
\floatname{listing}{Listing}
\title{DetAIL : A Tool to Automatically Detect and Analyze Drift In Language}
\author {
    Nishtha Madaan\textsuperscript{\rm 1, \rm 4},
    Adithya Manjunatha \textsuperscript{\rm 2} \thanks{Work done during Extreme Blue Internship at IBM},
    Hrithik Nambiar \textsuperscript{\rm 2}
    \footnotemark[1],
    Aviral Kumar Goel \textsuperscript{\rm 2} \footnotemark[1],\\
    Harivansh Kumar \textsuperscript{\rm 3},
    Diptikalyan Saha \textsuperscript{\rm 1},
    Srikanta Bedathur\textsuperscript{\rm 4}
}
\affiliations {
    \textsuperscript{\rm 1} IBM Research India\\
    \textsuperscript{\rm 2} Birla Institute of Technology and Science, Goa, India \\
    \textsuperscript{\rm 3} IBM Watson Openscale, India\\
    \textsuperscript{\rm 4} Indian Institute of Technology Delhi, India\\
    nishthamadaan@in.ibm.com,
    f20190118@goa.bits-pilani.ac.in, f20190100@goa.bits-pilani.ac.in, f20190166@goa.bits-pilani.ac.in,
    harivkum@in.ibm.com,
    diptsaha@in.ibm.com,
    srikanta@cse.iitd.ac.in \\
}

\usepackage{bibentry}

\begin{document}

\maketitle

\begin{abstract}
Machine learning and deep learning-based decision making has become part of today's software. The goal of this work is to ensure that machine learning and deep learning-based systems are as trusted as traditional software. Traditional software is made dependable by following rigorous practice like static analysis, testing, debugging, verifying, and repairing throughout the development and maintenance life-cycle. Similarly for machine learning systems, we need to keep these models up to date so that their performance is not compromised. For this, current systems rely on scheduled re-training of these models as new data kicks in. In this work, we propose to measure the data drift that takes place when new data kicks in so that one can adaptively re-train the models whenever re-training is actually required irrespective of schedules. In addition to that, we generate various explanations at sentence level and dataset level to capture why a given payload text has drifted. 

\end{abstract}

\section{Introduction}
\label{sec:intro}
Machine learning and deep learning-based decision making has become part of today's software. Traditional softwares are tested rigorously end-to-end to make sure that there is no undesired behavior. Similarly, Machine Learning and Deep Learning based solutions follow a similar paradigm by introduction of AI Testing strategies. 

These testing strategies encompass a set of testing strategies to make sure that machine learning model behaves in an expected way. One such important problem is detecting inconsistencies in text samples from production data. We call this text drift. This can be most commonly seen in a setting with temporal data and the data distribution changes over time. The idea is to track this distribution change as soon as possible and take necessary actions like re-training the model or generate more synthetic samples similar to drifted samples that can be used to re-train the model.

Given the fact that the problem to identify drifted data in text is key to health of a given model, just identifying drift is not enough. We should be able to point out what is going wrong with the new samples. Is it a totally different semantic space? Is there some syntactic structural differences? Does it have to do with few words which are unseen in the training distribution? Hence, in this paper we answer the above research questions and propose a system that can capture these aspects in text.

\begin{figure}[!h]
    \includegraphics[width=0.48\textwidth,height=0.2\textwidth]{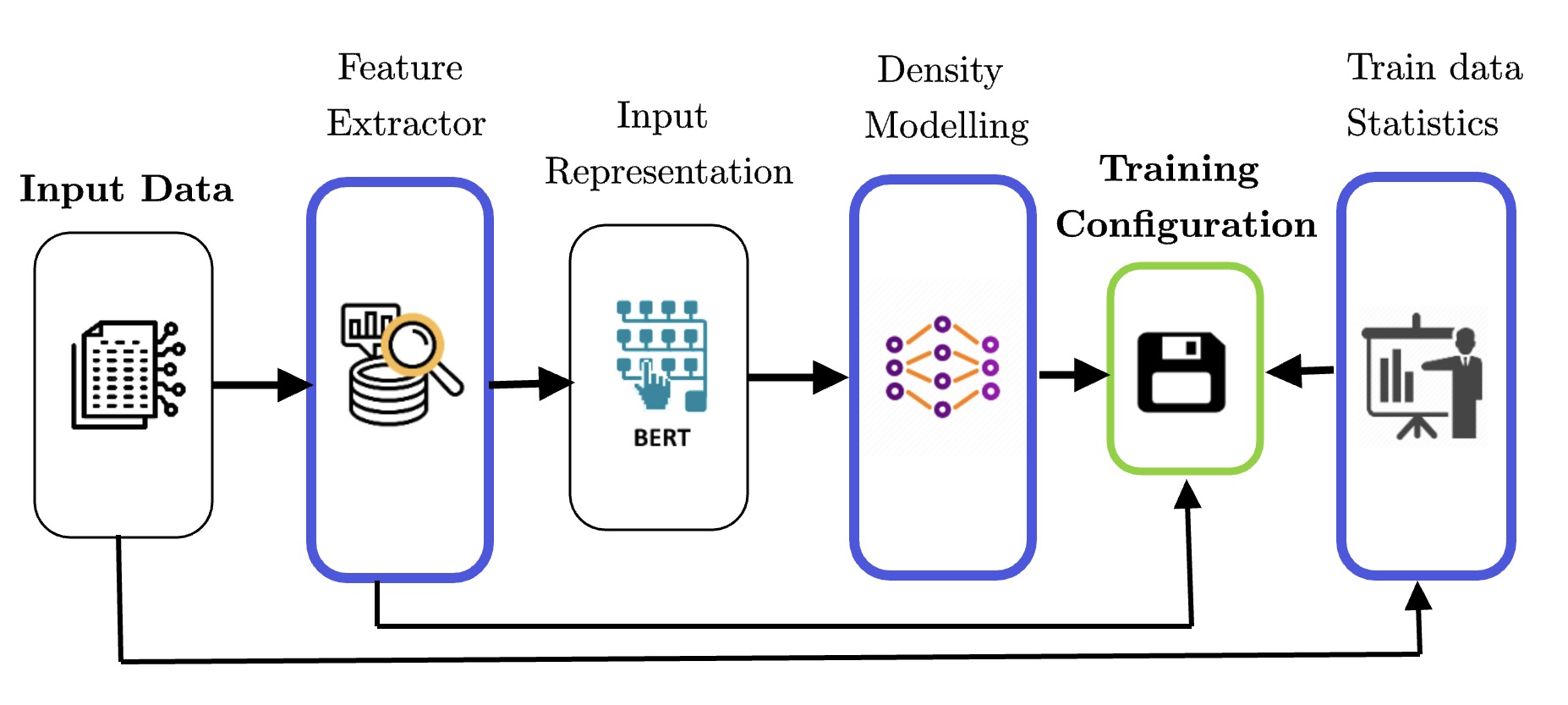}\\
    \includegraphics[width=0.48\textwidth,height=0.2\textwidth]{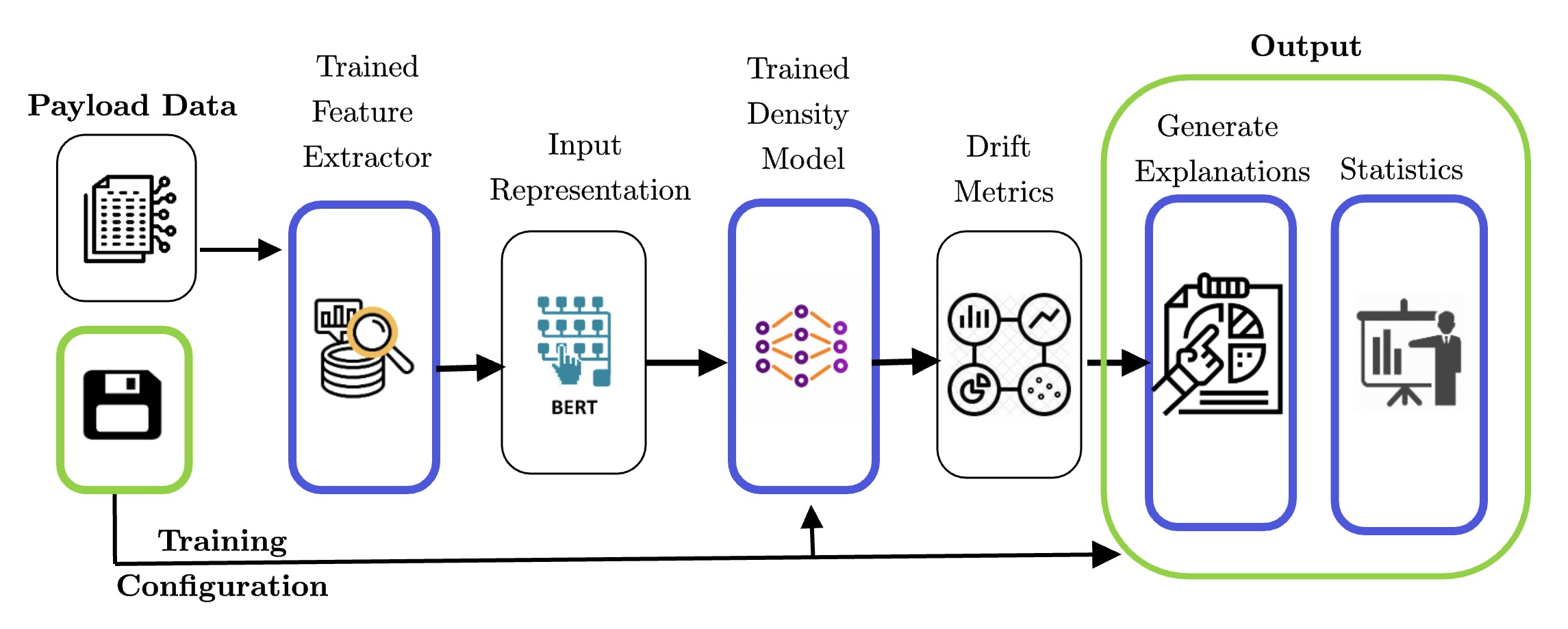}
    \caption{(a) Architecture of the training pipeline of our tool DetAIL. 
    (b) Architecture of the inference pipeline of our tool DetAIL.}
    \label{fig:arcg}
\end{figure}

Recent years have seen a tremendous interest in Out-Of-Distribution detection \citep{zheng2020out} \citep{arora2021types}. A recent work by \citep{feldhans2021drift} compares the performance of existing drift detectors on text classifiers based on high-dimensional document embeddings. Through their experiments they show that current drift detectors perform better on smaller embedding dimension and this serves as a motivation to develop drift detectors specifically tailored to text data.

Motivated by this work, we propose a tool \emph{DetAIL} to detect data drift in text data and generate explanations for that drift. Drift in text is defined as when a test sample exhibits different characteristics than the training sample set of a model. The main technical challenge while detecting drift in text is that text itself is very diverse. Hence it is very difficult to detect if a given text has drifted or has been rephrased and still similar to training data. In our tool, we build an input representation from training data and use different algorithms for density modelling to predict if a given payload sample is a drifted sample or not. In addition to this, we build an explanation module which generates explanations at sample level and dataset level to capture certain different characteristics exhibited by training data and payload data. A high-level view of our proposed architecture is described in Figure \ref{fig:arcg}.

\textbf{Contributions.} The main contributions of the paper can be seen as five fold : 1) We propose a configurable and an end-to-end system called \emph{DetAIL} to detect data drift in text, 2) We propose different algorithms to compute data drift and empirically show that which approach yields higher accuracy, 3) We propose a new metric to compare and contrast various algorithms used to identify such drift, 4) Our system generates sample level explanations to pin-point which parts in the text are responsible for drift, 5) We compute certain drift statistics which capture a dataset level view to compare syntactic differences in payload data and the train data.





\section{Components of Proposed System}
\label{sec:method}

\subsection{Feature Extractor Module}
In this section we discuss the different feature extraction methods we explored extract meaningful features. For all the approaches we use the training data to extract features. The input sentences are first pre-processed wherein any emojis, urls, and html tags were removed. We remove all the stopwords and lemmatize words for all word embedding based approaches. This is because sentence BERT, uses the entire sentence to generate embeddings and hence stop words have not been removed to preserve the context.

\textbf{Word Embeddings.} We train a Word2Vec model to generate embeddings for each word in the training data. The output is a 300 dimension vector, which is then averaged to generate document embedding. We compute the cosine-similarity between these embeddings of the training and payload data to yield the similarity score. We also compare the results with a pretrained Word2Vec model taken from Gensim \cite{rehurek_lrec} and a pre-trained GLoVe model. We observed better accuracy for drift detection using Word2Vec approach.


\textbf{BERT Word Embeddings.} We used a pretrained model bert-base-cased \cite{DBLP:journals/corr/abs-1810-04805} to generate tokens from an input sentence. The output is a 768 dimension vector for each token. The final vector is then generated by averaging these 768 dimension vectors across all token in a sentence.

\textbf{Sentence BERT Embeddings.} We used the "all-MiniLM-L6-v2" sentence-BERT (sBERT) pretrained model from Hugging Face \cite{reimers-2019-sentence-bert} to extract features from the input sentence. The output is a 384 dimensional vector for the entire sentence.

Out of all the embeddings, sBERT was fast to train and gave overall better results when used in conjunction with VAE based density modelling approach.

\subsection{Density Modelling Module}
In this section, we discuss the approaches we explored to model the distribution of training data. As a baseline, we used Word2Vec and GloVe based feature extractions with no density modelling.

\textbf{Gaussian Mixture Modelling: } We train a Gaussian Mixture Model on the BERT word embeddings extracted. We determine the \emph{n\_components}, that is, the number of clusters in the GMM by running experiments from \emph{n\_components} = 2 to 8 and determine silhoutte scores. The fit is best for the silhoutte score closest to 1. Using \emph{n\_components} value, the GMM is trained. We generate log likelihood of the payload data to lie on the trained GMM. This score is then dividing by the dimension of embeddings (768 for BERT) and then scaled to a scale of 0 to 1, using Min-Max scaling.

\textbf{Variational Auto Encoder: } We train a VAE on top of the sBERT embeddings generated in the previous module. The VAE loss used here is the sum of Reconstruction Loss and KL Divergence Loss which is also known as ELBO (evidence lower bound) Loss. During inference, this loss is used to calculate the drift score. The general idea behind this method is that if the VAE is able reconstruct the payload data accurately then the payload data is not drifted. The output is in range of 0 to infinity which is scaled from 0 to 1 using negative exponentiation (e$^{-x}$), to yield the similarity score.

\subsection{Explanations}
\subsubsection{Sample Level Explanations}
In this section, we describe our method for explaining the drift prediction given an utterance x and it’s drift score s. We implement this using word masking. Word masking is a method which involves hiding certain words from the input sentence. For every word i in x, we check the drift score upon masking i in x, using the change in the score caused by this masking we determine the contribution of each word towards the drift score. On masking a word i, if the score, $s_i$ ,  shifts away from being drifted by $h_i$ , we can say that i was contributing to drift by $h_i$. Doing this for all words in x we determine the contribution of each word towards the drift.

$ h_i = s_i - s$
\begin{figure}[!h]
    \centering
    \includegraphics[width=0.4\textwidth]{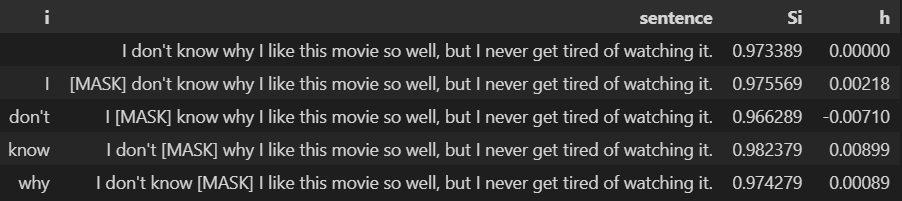}
    \includegraphics[width=0.47\textwidth]{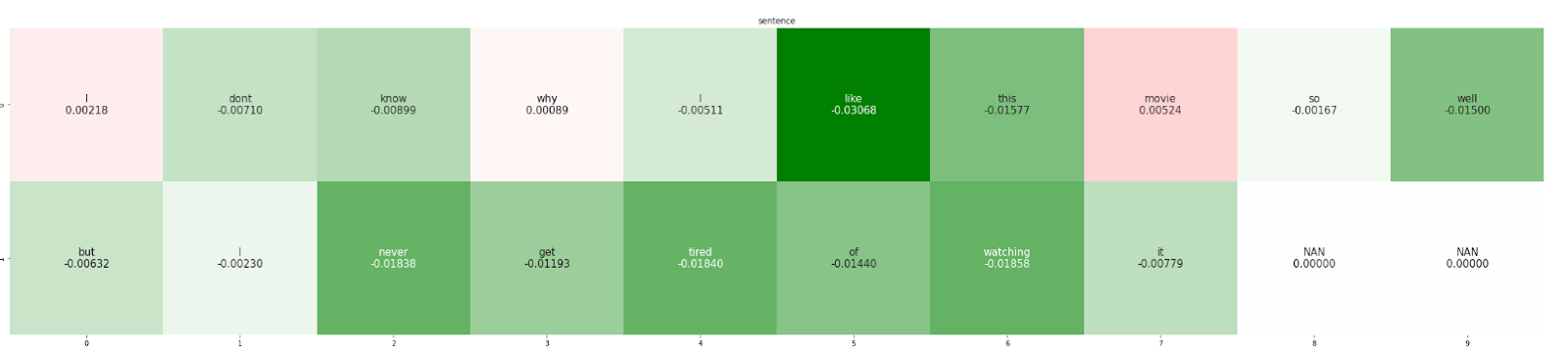}
    \caption{(a) Illustration of word masking. (b) Output of word masking approach shows which word contributes to maximum drift.}
    \label{fig:masking}
\end{figure}

\subsubsection{Dataset Level Explanations}
We compute the following dataset-level explanations :\\
\textbf{1. Verb-neighbourhood patterns: } In the english language verbs are one of the most important parts of the sentence, most if not all sentences are centered around verbs. Verb-neighbourhood patterns uses this property of english sentences to capture patterns in the dataset. We first use Regex-based chunker to chunk the Parts-Of-Speech (POS) tags of the sentences into Noun Phrases(NP) and other tags, in order to reduce the number of tags that we are working with.

\begin{figure}[!h]
    \centering
    \includegraphics[width=0.45\textwidth]{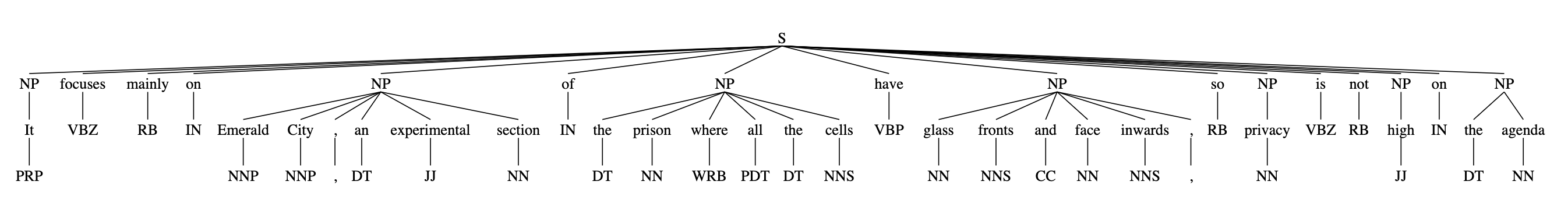}
    \caption{Regex based Noun Phrase chunking.}
    \label{fig:regex_chunk}
\end{figure}

Finally, we capture the patterns in the bounded region of 2 tags before and after the verb tag (Verb - Neighbourhood). We then calculate the frequencies and hence probabilities of occurrence of these Verb - Neighbourhood patterns in the dataset.

We repeat this process for the payload dataset as well and compare the probabilities of the common patterns. We also report the new patterns found in the payload dataset.

\begin{figure}[!h]
    \includegraphics[width=0.35\textwidth,height=0.2\textwidth]{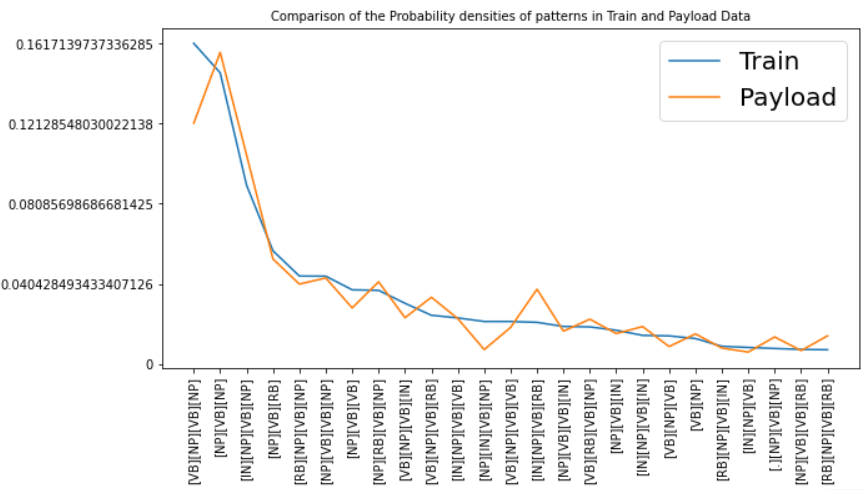}
    \includegraphics[width=0.35\textwidth,height=0.2\textwidth]{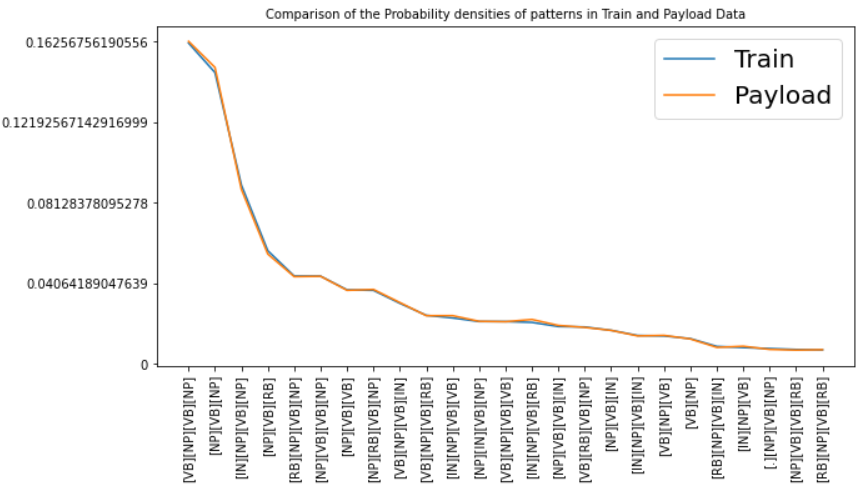}
    \caption{Comparison of the probabilities of the top 25 verb neighbourhood patterns in Train (Insurance Company Reviews) and Payload data for,(a) drifted payload (Fashion Reviews). (b) non-drifted payload (Insurance Reviews Held-out)}
    \label{fig:verb_rule_stats}
\end{figure}

\begin{table*}[t]
\begin{tabular}{|l|l|ll|ll|ll|}
\hline
\multirow{2}{*}{{\ul \textbf{Train Dataset}}}     & \multirow{2}{*}{{\ul \textbf{Payload Dataset}}} & \multicolumn{2}{l|}{{\ul \textbf{Word2Vec}}} & \multicolumn{2}{l|}{{\ul \textbf{BERT + GMM}}}    & \multicolumn{2}{l|}{{\ul \textbf{sBERT + VAE}}}    \\ \cline{3-8} 
                                                  &                                                 & \multicolumn{1}{l|}{Accuracy $\uparrow$}   & Time $\downarrow$      & \multicolumn{1}{l|}{Accuracy $\uparrow$}           & Time  $\downarrow$  & \multicolumn{1}{l|}{Accuracy $\uparrow$}           & Time  $\downarrow$   \\ \hline
\multirow{5}{*}{{\ul \textbf{IMDB}}}              & AGNews                                          & \multicolumn{1}{l|}{94.815 \%}  & 0.8 min    & \multicolumn{1}{l|}{\textbf{99.765 \%}} & 60 mins & \multicolumn{1}{l|}{99.228 \%}          & 2.5 mins \\ \cline{2-8} 
                                                  & YELP                                            & \multicolumn{1}{l|}{98.975 \%}  & 0.05 min   & \multicolumn{1}{l|}{\textbf{99.868 \%}} & 1 min   & \multicolumn{1}{l|}{99.062 \%}          & 0.5 min  \\ \cline{2-8} 
                                                  & Fashion Reviews                                 & \multicolumn{1}{l|}{76.521 \%}  & 0.17 min   & \multicolumn{1}{l|}{\textbf{99.929 \%}} & 15 mins & \multicolumn{1}{l|}{99.906 \%}          & 2.5 mins \\ \cline{2-8} 
                                                  & Restaurant Reviews                              & \multicolumn{1}{l|}{96.812 \%}  & 0.017 min  & \multicolumn{1}{l|}{\textbf{99.568 \%}} & 0.5 min & \multicolumn{1}{l|}{94.800 \%}          & 0.05 min \\ \cline{2-8} 
                                                  & Insurance Reviews                               & \multicolumn{1}{l|}{85.359 \%}  & 0.34 min   & \multicolumn{1}{l|}{87.711 \%}          & 30 mins & \multicolumn{1}{l|}{\textbf{99.136 \%}} & 5 mins   \\ \hline
\multirow{5}{*}{{\ul \textbf{Insurance Reviews}}} & AGNews                                          & \multicolumn{1}{l|}{61.852 \%}  & 0.5 min    & \multicolumn{1}{l|}{\textbf{96.487 \%}} & 60 mins & \multicolumn{1}{l|}{94.706 \%}          & 2.5 mins \\ \cline{2-8} 
                                                  & YELP                                            & \multicolumn{1}{l|}{58.308 \%}  & 0.1 min    & \multicolumn{1}{l|}{98.386 \%}          & 1 min   & \multicolumn{1}{l|}{\textbf{98.960 \%}} & 0.5 min  \\ \cline{2-8} 
                                                  & Fashion Reviews                                 & \multicolumn{1}{l|}{60.728 \%}  & 0.2 min    & \multicolumn{1}{l|}{\textbf{98.376 \%}} & 15 mins & \multicolumn{1}{l|}{96.182 \%}          & 2.5 mins \\ \cline{2-8} 
                                                  & Restaurant Reviews                              & \multicolumn{1}{l|}{69.929 \%}  & 0.017 min  & \multicolumn{1}{l|}{80.636 \%}          & 0.5 min & \multicolumn{1}{l|}{\textbf{81.574 \%}} & 0.05 min \\ \cline{2-8} 
                                                  & IMDB                                            & \multicolumn{1}{l|}{58.198 \%}  & 0.85       & \multicolumn{1}{l|}{50.261\%}           & 36 mins & \multicolumn{1}{l|}{\textbf{98.633 \%}} & 24 mins  \\ \hline
\end{tabular}
\caption{Comparison of accuracy and inference times of the three above mentioned methods.}
\label{tab:accuracy}
\end{table*}

\begin{figure}[!h]
    \includegraphics[width=0.48\textwidth,height=0.3\textwidth]{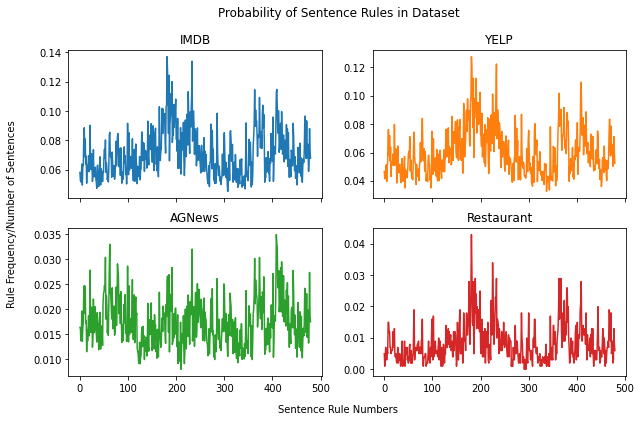}
    \caption{Comparison of the probabilities of occurrence of Sentence Rules in different datasets. This statistic helps to correctly infer that YELP and IMDB Datasets are structurally very similar.}
    \label{fig:Sentence_rules_stats}
\end{figure}

\textbf{2. Sentence Rules: } In conjunction with capturing changes in context leading to drift, it is important to detect changes in the structure of the sentences. We utilize Part-Of-Speech (POS) tags to identify patterns present in a sentence of the dataset. First, we tag every sentence in the dataset using the POS tags of their constituent words. We search through all possible rules which could occur in these POS-tagged sentences. These rules are combinations of six important POS tags - Nouns, Pronouns, Verb, Adverb, Adjective, and Determiner. The rules are designed as regular expressions with each tag appearing exactly once. There are 480 such rules which amounts to all common combinations of the six POS tags. We then search through the dataset to capture the probability of a particular rule occurring in a sentence of the dataset. 
These probabilities are used further to identify the commonly occurring and newly found sentence rules in a dataset. The newly found rules are the rules which had a very low probability of occurrence in the training data but is commonly seen in the payload data. We observe that structurally similar datasets often show similar rule probability distributions as shown in Figure ~\ref{fig:Sentence_rules_stats}.

\textbf{3. Drift due to Named Entities: } We use Stanford Dependency Parser \cite{spacy2} to perform Named Entity Recognition on the entire dataset and draw inferences using the frequency of occurrence of various NER tags. For instance, we can infer that the payload data has a lot more organisation names as compared to training data which might result in drift.

\textbf{4. Drift due to dependencies: } We identify the frequency of various dependency tags to draw inferences. For instance, we can infer that the payload data is 2 percent more passive than the training data (based on the frequency of \emph{nsubjpass} dependency tag across both the datasets).

\textbf{5. Relationship between NER and DEP: } Extrapolating the results from previous two statistics we determine a relationship between NER tags and their dependencies in the dataset which helps us identify the context drift due to NER tags across the datasets. We extract the top two predominant dependencies for the specific NER tags across the training data and payload data as shown in Fig 4.

\begin{figure}[!h]
    \centering
    \includegraphics[width=0.45\textwidth]{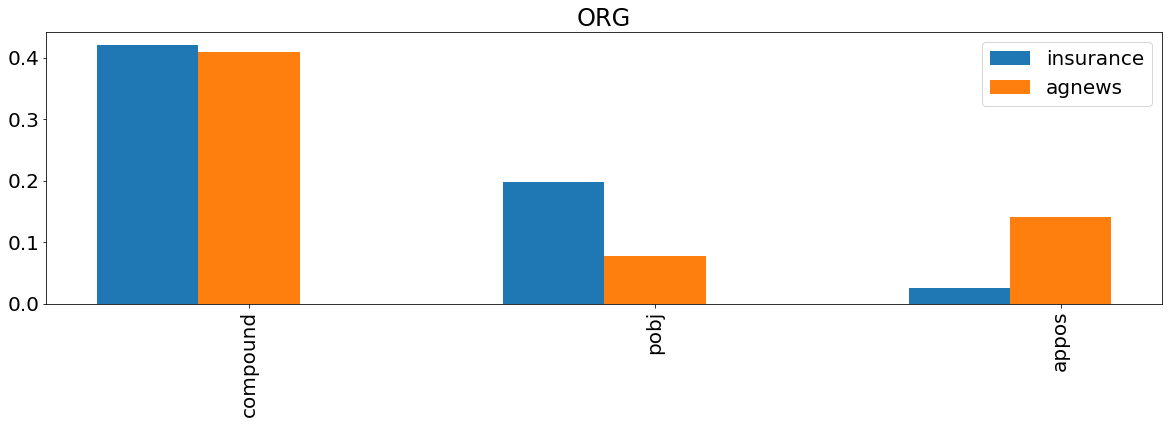}
    \caption{The top two dependencies for the ORG NER tag in Insurance dataset are compound and pobj whereas in the agnews dataset are compound and appos.}
    \label{fig:NER_stats}
\end{figure}

\textbf{6. Drift due to Noun and Verb Phrase Chunks: } We train a tag chunker on the CoNLL-2000 dataset \cite{tksbuchholz2000conll} and use it to perform noun and verb phrase chunking on the training and payload data. This helps us explain drift in terms of chunk phrases. For instance, we can infer that "There are five verb phrases per sentence in the Insurance Reviews per sentence as compared to three verb phrases per sentence in the agnews dataset". 

\begin{table*}[!h]
\small
\begin{tabular}{@{}|p{0.95\textwidth}|p{0.45\textwidth}|@{}}


\hline
\\
\textbf{Input Representation} : SBERT\\
\textbf{Density Model} : VAE\\
\textbf{Training Dataset}: Insurance Company Reviews\\
\textbf{Payload Dataset}: AGNews\\
\textbf{Payload Sentence}: \textit{If you can't get a good night's sleep it's likely that your parents are at least partly to blame.} \\
\textbf{Threshold} : \textit{0.995}\\ 
\textbf{Output from Trained Drift Predictor, Similarity Score} : Drifted, 0.8814285151404778\\

\textbf{Sample Level Explanations for Drift} :\\ 

\includegraphics[width=0.9\textwidth, height=20mm]{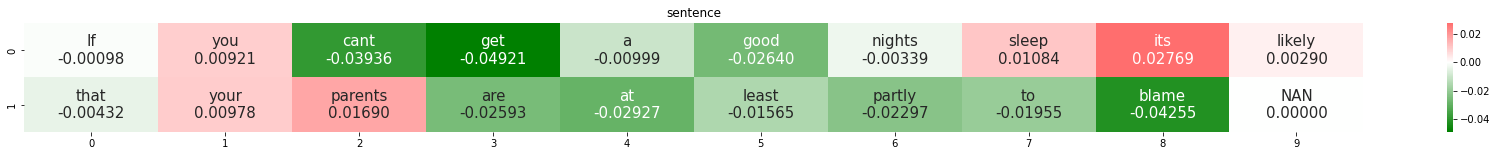}
\\
\textbf{Drift Statistics} : \\
1) \textbf{Verb neighborhood Patterns} : 
\begin{table}[H]
\centering
\begin{tabular}{|l|l|l|} 
\hline
\textbf{Pattern}                                                                                    & \textbf{Example}                                                                                                                                                                                                                & \textbf{Likelihood Percentage in Train}  \\ 
\hline
\textcolor{blue}{[VB]}\textcolor[rgb]{0,0.502,0}{[NP]}\textcolor{yellow}{[VB]}\textcolor{red}{[NP]} & \begin{tabular}[c]{@{}l@{}}If you ca n't~\textcolor{blue}{get~}\textcolor[rgb]{0,0.502,0}{a good night 's sleep it~}\textcolor{yellow}{'s~}\textcolor{red}{likely~}that your \\parents are at least partly to blame\end{tabular} & 16.1714 \%                       \\
\hline
\textcolor{blue}{[NP]}\textcolor[rgb]{0,0.502,0}{[RB]}\textcolor{yellow}{[VB]}\textcolor{red}{[NP]} & \begin{tabular}[c]{@{}l@{}}\textcolor{blue}{If you ca~}\textcolor[rgb]{0,0.502,0}{n't~}\textcolor{yellow}{get~}\textcolor{red}{a good night 's sleep it~}'s likely that your \\parents are at least partly to blame\end{tabular} & 3.7162 \%                       \\ 
\hline
\textcolor{blue}{[IN]}\textcolor[rgb]{0,0.502,0}{[NP]}\textcolor{yellow}{[VB]}\textcolor{red}{[IN]} & If you ca n't get a good night 's sleep it 's likely~\textcolor{blue}{that~}&\\&\textcolor[rgb]{0,0.502,0}{your parents} \textcolor{yellow}{are~}\textcolor{red}{at~}least partly to blame                                           & 1.4453 \%                      \\ 

\hline
\textcolor{blue}{[RB]}\textcolor[rgb]{0,0.502,0}{[NP]}\textcolor{yellow}{[VB]}                      & \begin{tabular}[c]{@{}l@{}}If you ca n't get a good night 's sleep it 's likely that your \\parents are at least~\textcolor{blue}{partly~}\textcolor[rgb]{0,0.502,0}{to~}\textcolor{yellow}{blame}\end{tabular}                  & 0.3726 \%                       \\ 

\hline
\end{tabular}
\end{table}

2) \textbf{Sentence Rule} :

\begin{table}[H]
\centering
\begin{tabular}{|l|l|} 
\hline
\textbf{New Sentence Rules}                                                                                                                                                                                                                                                                                                                                                                                                                                                                                                                                                                                                                                          & \textbf{Sentence}                                                                                                                                                                                                                                                                                                                                                                                                                                                                                                                                                                                                        \\ 
\hline
\begin{tabular}[c]{@{}l@{}}\textcolor[rgb]{0.129,0.129,0.129}{\#415: (}\textcolor[rgb]{0,0.392,0}{DET}\textcolor[rgb]{0.129,0.129,0.129}{)+\textbackslash{}S+(}\textcolor[rgb]{0,0.392,0}{ADJ}\textcolor[rgb]{0.129,0.129,0.129}{)+\textbackslash{}S+(}\textcolor[rgb]{0,0.392,0}{PRON}\textcolor[rgb]{0.129,0.129,0.129}{)+\textbackslash{}S+}\\\textcolor[rgb]{0.129,0.129,0.129}{~ ~ ~ ~ ~ (}\textcolor[rgb]{0,0.392,0}{NOUN}\textcolor[rgb]{0.129,0.129,0.129}{)+\textbackslash{}S+(}\textcolor[rgb]{0,0.392,0}{ADV}\textcolor[rgb]{0.129,0.129,0.129}{)+\textbackslash{}S+(}\textcolor[rgb]{0,0.392,0}{VERB}\textcolor[rgb]{0.129,0.129,0.129}{)+}\end{tabular} & \begin{tabular}[c]{@{}l@{}}\textcolor[rgb]{0.129,0.129,0.129}{If you ca n't get }\textcolor[rgb]{0,0.392,0}{a }\textcolor[rgb]{0.129,0.129,0.129}{ }\textcolor[rgb]{0,0.392,0}{good }\textcolor[rgb]{0.129,0.129,0.129}{ night 's sleep }\textcolor[rgb]{0,0.392,0}{it}\textcolor[rgb]{0.129,0.129,0.129}{ 's likely that your }\\\textcolor[rgb]{0,0.392,0}{parents }\textcolor[rgb]{0.129,0.129,0.129}{ are at least }\textcolor[rgb]{0,0.392,0}{partly }\textcolor[rgb]{0.129,0.129,0.129}{ to }\textcolor[rgb]{0,0.392,0}{blame}\textcolor[rgb]{0.129,0.129,0.129}{.}                                         \end{tabular}                                      \\ 
\hline
\begin{tabular}[c]{@{}l@{}}\textcolor[rgb]{0.129,0.129,0.129}{\#180: (}\textcolor[rgb]{0,0.392,0}{PRON}\textcolor[rgb]{0.129,0.129,0.129}{)+\textbackslash{}S+(}\textcolor[rgb]{0,0.392,0}{VERB}\textcolor[rgb]{0.129,0.129,0.129}{)+\textbackslash{}S+(}\textcolor[rgb]{0,0.392,0}{DET}\textcolor[rgb]{0.129,0.129,0.129}{)+\textbackslash{}S+}\\\textcolor[rgb]{0.129,0.129,0.129}{~ ~ ~ ~ ~ (}\textcolor[rgb]{0,0.392,0}{NOUN}\textcolor[rgb]{0.129,0.129,0.129}{)+\textbackslash{}S+(}\textcolor[rgb]{0,0.392,0}{ADJ}\textcolor[rgb]{0.129,0.129,0.129}{)+\textbackslash{}S+(}\textcolor[rgb]{0,0.392,0}{ADV}\textcolor[rgb]{0.129,0.129,0.129}{)+}\end{tabular} & \begin{tabular}[c]{@{}l@{}}\textcolor[rgb]{0.129,0.129,0.129}{If }\textcolor[rgb]{0,0.392,0}{you }\textcolor[rgb]{0.129,0.129,0.129}{ ca n't }\textcolor[rgb]{0,0.392,0}{get }\textcolor[rgb]{0.129,0.129,0.129}{ }\textcolor[rgb]{0,0.392,0}{a }\textcolor[rgb]{0.129,0.129,0.129}{good }\textcolor[rgb]{0,0.392,0}{night}\textcolor[rgb]{0.129,0.129,0.129}{ 's sleep it 's }\textcolor[rgb]{0,0.392,0}{likely }\textcolor[rgb]{0.129,0.129,0.129}{ that your }\\\textcolor[rgb]{0.129,0.129,0.129}{parents are at least }\textcolor[rgb]{0,0.392,0}{partly }\textcolor[rgb]{0.129,0.129,0.129}{ to blame .}\end{tabular}  \\ 
\hline
\begin{tabular}[c]{@{}l@{}}\textcolor[rgb]{0.129,0.129,0.129}{\#385: (}\textcolor[rgb]{0,0.392,0}{DET}\textcolor[rgb]{0.129,0.129,0.129}{)+\textbackslash{}S+(}\textcolor[rgb]{0,0.392,0}{PRON}\textcolor[rgb]{0.129,0.129,0.129}{)+\textbackslash{}S+(}\textcolor[rgb]{0,0.392,0}{NOUN}\textcolor[rgb]{0.129,0.129,0.129}{)+\textbackslash{}S+}\\\textcolor[rgb]{0.129,0.129,0.129}{~ ~ ~ ~ ~ (}\textcolor[rgb]{0,0.392,0}{ADJ}\textcolor[rgb]{0.129,0.129,0.129}{)+\textbackslash{}S+(}\textcolor[rgb]{0,0.392,0}{ADV}\textcolor[rgb]{0.129,0.129,0.129}{)+\textbackslash{}S+(}\textcolor[rgb]{0,0.392,0}{VERB}\textcolor[rgb]{0.129,0.129,0.129}{)+}\end{tabular} & \begin{tabular}[c]{@{}l@{}}\textcolor[rgb]{0.129,0.129,0.129}{If you ca n't get }\textcolor[rgb]{0,0.392,0}{a }\textcolor[rgb]{0.129,0.129,0.129}{ good night 's sleep }\textcolor[rgb]{0,0.392,0}{it}\textcolor[rgb]{0.129,0.129,0.129}{ 's likely that your }\\\textcolor[rgb]{0,0.392,0}{parents }\textcolor[rgb]{0.129,0.129,0.129}{ are at }\textcolor[rgb]{0,0.392,0}{least }\textcolor[rgb]{0.129,0.129,0.129}{ }\textcolor[rgb]{0,0.392,0}{partly }\textcolor[rgb]{0.129,0.129,0.129}{ to }\textcolor[rgb]{0,0.392,0}{blame}\textcolor[rgb]{0.129,0.129,0.129}{ .}                                        \end{tabular}                                       \\ 
\hline
\begin{tabular}[c]{@{}l@{}}\textcolor[rgb]{0.129,0.129,0.129}{\#361: (}\textcolor[rgb]{0,0.392,0}{DET})\textcolor[rgb]{0.129,0.129,0.129}{+\textbackslash{}S+(}\textcolor[rgb]{0,0.392,0}{NOUN}\textcolor[rgb]{0.129,0.129,0.129}{)+\textbackslash{}S+(}\textcolor[rgb]{0,0.392,0}{PRON}\textcolor[rgb]{0.129,0.129,0.129}{)+\textbackslash{}S+}\\\textcolor[rgb]{0.129,0.129,0.129}{~ ~ ~ ~ ~ (}\textcolor[rgb]{0,0.392,0}{ADJ}\textcolor[rgb]{0.129,0.129,0.129}{)+\textbackslash{}S+(}\textcolor[rgb]{0,0.392,0}{ADV}\textcolor[rgb]{0.129,0.129,0.129}{)+\textbackslash{}S+(}\textcolor[rgb]{0,0.392,0}{VERB}\textcolor[rgb]{0.129,0.129,0.129}{)+}\end{tabular} & \begin{tabular}[c]{@{}l@{}}\textcolor[rgb]{0.129,0.129,0.129}{If you ca n't get }\textcolor[rgb]{0,0.392,0}{a }\textcolor[rgb]{0.129,0.129,0.129}{ good }\textcolor[rgb]{0,0.392,0}{night}\textcolor[rgb]{0.129,0.129,0.129}{ 's sleep }\textcolor[rgb]{0,0.392,0}{it}\textcolor[rgb]{0.129,0.129,0.129}{ 's }\textcolor[rgb]{0,0.392,0}{likely }\textcolor[rgb]{0.129,0.129,0.129}{ that your }\\\textcolor[rgb]{0.129,0.129,0.129}{parents are at least }\textcolor[rgb]{0,0.392,0}{partly }\textcolor[rgb]{0.129,0.129,0.129}{ to }\textcolor[rgb]{0,0.392,0}{blame}\textcolor[rgb]{0.129,0.129,0.129}{.}\end{tabular}  \\ 
\hline
\begin{tabular}[c]{@{}l@{}}\textcolor[rgb]{0.129,0.129,0.129}{\#182: (}\textcolor[rgb]{0,0.392,0}{PRON}\textcolor[rgb]{0.129,0.129,0.129}{)+\textbackslash{}S+(}\textcolor[rgb]{0,0.392,0}{VERB}\textcolor[rgb]{0.129,0.129,0.129}{)+\textbackslash{}S+(}\textcolor[rgb]{0,0.392,0}{DET}\textcolor[rgb]{0.129,0.129,0.129}{)+\textbackslash{}S+}\\\textcolor[rgb]{0.129,0.129,0.129}{~ ~ ~ ~ ~ (}\textcolor[rgb]{0,0.392,0}{ADJ}\textcolor[rgb]{0.129,0.129,0.129}{)+\textbackslash{}S+(}\textcolor[rgb]{0,0.392,0}{NOUN}\textcolor[rgb]{0.129,0.129,0.129}{)+\textbackslash{}S+(}\textcolor[rgb]{0,0.392,0}{ADV}\textcolor[rgb]{0.129,0.129,0.129}{)+}\end{tabular} & \begin{tabular}[c]{@{}l@{}}\textcolor[rgb]{0.129,0.129,0.129}{If }\textcolor[rgb]{0,0.392,0}{you }\textcolor[rgb]{0.129,0.129,0.129}{ ca n't }\textcolor[rgb]{0,0.392,0}{get }\textcolor[rgb]{0.129,0.129,0.129}{ }\textcolor[rgb]{0,0.392,0}{a }\textcolor[rgb]{0.129,0.129,0.129}{ }\textcolor[rgb]{0,0.392,0}{good }\textcolor[rgb]{0.129,0.129,0.129}{ }\textcolor[rgb]{0,0.392,0}{night}\textcolor[rgb]{0.129,0.129,0.129}{ 's sleep it 's likely that your }\\\textcolor[rgb]{0.129,0.129,0.129}{parents are at least }\textcolor[rgb]{0,0.392,0}{partly }\textcolor[rgb]{0.129,0.129,0.129}{ to blame.}\end{tabular}  \\
\hline
\end{tabular}
\end{table}                                                                                 

3) \textbf{Dependency of particular NER tag in sample vs dataset}:
\begin{table}[H]
\centering
\begin{tabular}{|c|c|c|c|} 
\hline
\textbf{NER Tag} & \textbf{NER}            & \textbf{Dependency in sample} & \textbf{Top two most common dependencies in training dataset}  \\ 
\hline
TIME             & \textit{a good night's} & det                           & {[}pobj, nummod]                                      \\
\hline
\end{tabular}
\end{table}
\\
\hline
\end{tabular}
\caption{Illustration of the result generated by the proposed drift detection framework. The training data used is the Insurance Company Reviews dataset and the drifted payload sample is from the AGNews dataset. The frameworks accurately detects the drift and explains the words responsible for the drift along with some valuable statistics related to the structure of the payload sample.}
\label{tab:walkthrough}

\end{table*}

\section{Experimental Setup}
\label{sec:experiments}
The goal of the experiments is to \emph{evaluate the ability of our approach to detect drifted texts effectively.} We list the datasets details used in our evaluation below.

\subsection{Datasets}

\textbf{IMDB Reviews Dataset}\cite{maas2011learning} This dataset 50K movies reviews from IMDB. For pre-processing, we remove emoji’s and html tags.\\
\textbf{Yelp.} \cite{shen2017style} This dataset focuses on informal movie reviews from YELP. It contains 1.7K movie reviews.\\
\textbf{Fashion reviews} \cite{fashion_dataset} This dataset focuses on various reviews of clothing by customers on a Women's clothing E-Commerce website. It contains 23K reviews.\\
\textbf{Restaurant reviews.} \cite{restaurant_reviews} This dataset focuses on reviews of restaurants from various customers. It contains 1K reviews.\\
\textbf{AgNews.}\cite{agnews} This dataset contains real world news articles from the web, belonging to 4 major classes - business, sports, sci-fi and world categories. The entire dataset contains 120K training samples and 7.6K testing samples.\\
\textbf{Insurance Company reviews.} \cite{insurance_reviews} This dataset contains the reviews of a variety of customers around the world concerning insurance companies. It consists of 44.9K reviews.


\subsection{Metric for Evaluation}

Stratified Accuracy:- Stratified Accuracy is an accuracy-based metric which is used to score the correctness of the drift detection model. We divide the In-distribution dataset into train and held-out in a stratified manner. That is, if the In-distribution dataset has N intent classes, for each intent class we will have a set of utterances $D_i$ . We divide each $D_i$ into ($D_i^{ train}, D_i^{held-out}$) using a common split percentage. 
Then we use the $D_i^{train} $ of all the intent classes, i, for density modelling. We then define $D_{iid}$ to be the set of all the $D_i^{ held-out }$ from all the intent classes i, and $D_{ood}$ to be the set of all the utterances from the Drifted dataset. We report the scaled version of the accuracy to make sure that both drifted and non-drifted samples are accurately detected by our algorithm. The scaled accuracy is calculated as,

\[
Accuracy = \frac{\frac{(\#correct_{iid})*\#D_{ood}}{\#D_{iid }} +(\#correct_{ood}) }{2*\#( D_{ood})}
\]

\section{DetAIL : System Walk-Through}
In Table \ref{tab:walkthrough}, we show how different components of our system operate on a sample from the payload data. Due to confidentiality, we can not share a sentence from client data and hence show examples from opensource datasets. Our framework works in the following steps : \\
\textbf{Step 1. Drift Detection.}
In this case, we have trained a VAE using the sBERT embeddings of the Insurance Company Reviews dataset. The payload sentence is a sentence selected from the AGNews dataset, to test our framework. We use the inference pipeline shown in Figure \ref{fig:arcg} to detect if a given payload sample is drifted or not with a similarity score. In this case, we see that the payload text is drifted with 0.881 Similarity score. \\
\textbf{Step 2. Generating Sample Level Explanations.}
The next step is to generate explanations at a sample level. These explanations help us in interpreting output of our drift detection model and help us to understand why a given payload sample has been marked \emph{drifted}. This gives us insights on which words are responsible for causing the drift. As shown in the example, words like \texttt{parents}, \texttt{sleep} and other words marked in red are responsible for this drift. \\
\textbf{Step 3. Generating Drift Statistics.} 
In this step, we generate drift statistics. These statistics capture dataset-level differences in train data and the payload data. The first statistic compared \emph{Verb neighborhood Patterns.} We find out the patterns in the train dataset and those in the payload dataset and show that what is the likelihood percentage of a given pattern in the train dataset. This helps us in evaluating if a given payload sample shows a very different pattern than what the train dataset exhibits. For instance, in this case, \emph{[RB][NP][VB]} has 0.37\% likelihood percentage to be seen in train dataset and hence this is a new pattern seen in payload sample. The next statistic is on \emph{sentence rule} in which we generate POS-tag based patterns on training data and payload sample for each sentence. Then, we compare that if payload sample exhibits a new pattern that is not found commonly in train data. The patterns shown in Table \ref{tab:walkthrough} shows the patterns that have not been seen frequently in train data. The next statistic is on \emph{dependency of a NER}. In this statistic, we see if a given NER tag takes a certain dependency tag in train and completely different dependency tag in payload sample. In this example, we see that \texttt{a good night's} is an NER which takes \emph{det} in payload sample but in train dataset it takes \emph{pmod, nummod} dependency tags.

\section{Deployed Solution}
\label{sec:deployed}
\begin{figure}[!h]
    \includegraphics[width=0.48\textwidth,height=0.26\textwidth]{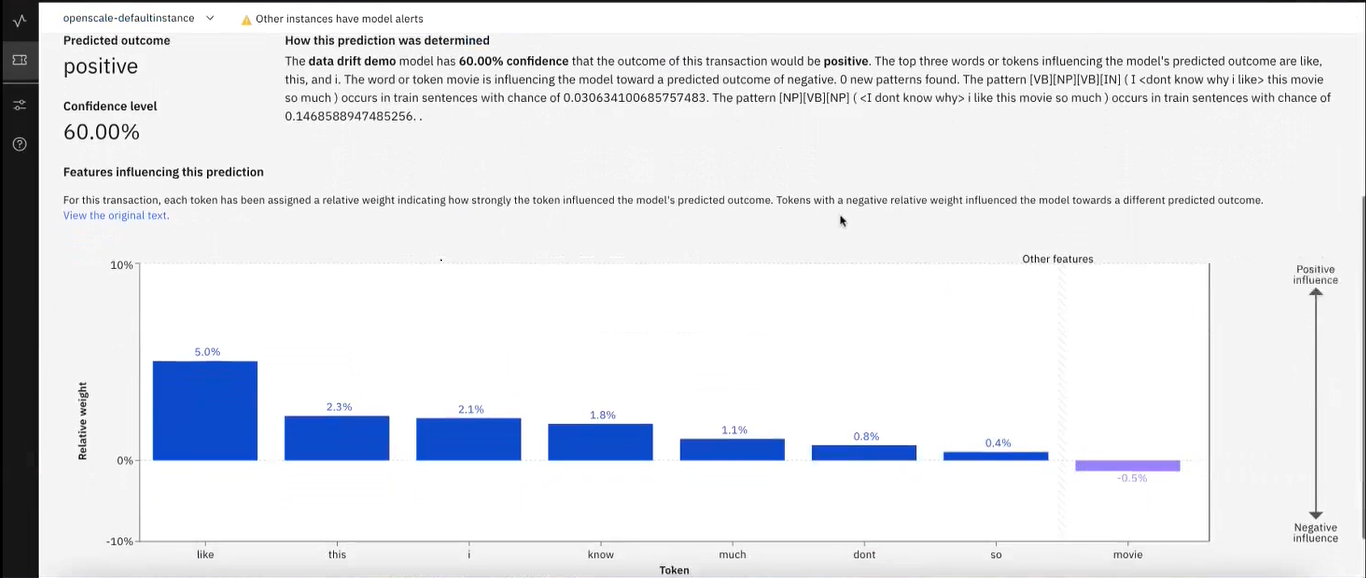}
    \includegraphics[width=0.48\textwidth,height=0.26\textwidth]{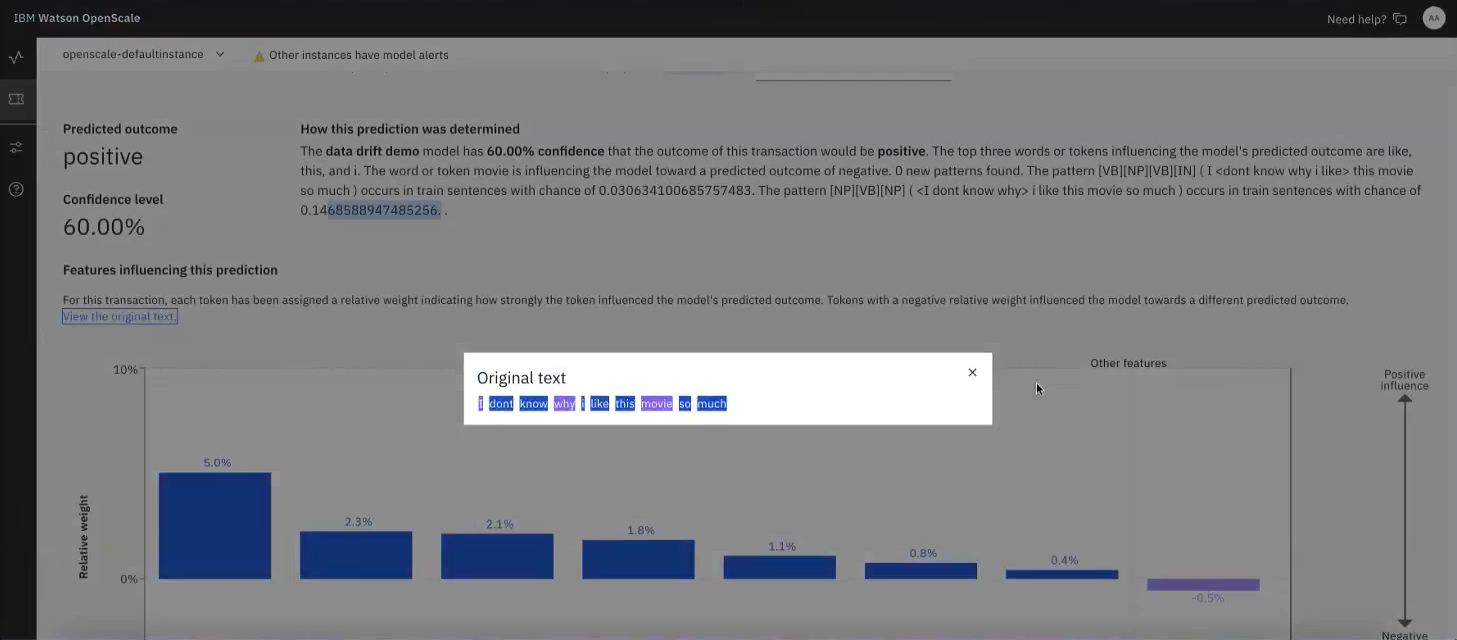}
    \caption{Deployed Text Drift Solution on Watson OpenScale.}
    \label{fig:openscale}
\end{figure}

We integrate this service into IBM Watson OpenScale which is an enterprise-grade environment for AI applications that helps to monitor deployed models. Users can build and train their text-based models in Watson Studio on the training data and save the required configurations. The trained model is deployed using Watson Machine Learning and this deployment is monitored real-time for data drift in Watson OpenScale using the proposed solution. Here, we have deployed a sentiment classifier trained on Insurance Company Reviews. We monitor this model in OpenScale for different payload sentences as shown in Figure \ref{fig:openscale}. 
Since we had best results (with respect to accuracy and inference time) using Sentence BERT as feature extractor and Variational Auto Encoder for density modelling, we deploy this approach for drift detection. We have used a user-defined threshold of 0.995 on the similarity score. Below this threshold score, the payload sample is identified to be drifted. Also, one can view which words in a given payload text contribute to drift. In addition to this, we have added a text explanation block which says \emph{How this prediction was determined?}. This block contains outputs for drift statistics representing differences in syntactical structure of payload text and training texts.

\section{Related Work}
\label{sec:related_work}
There exists substantial literature in concept drift on structured data. But the work on unstructured data including text is very limited. Greczo et al \cite{greco2021drift} propose Drift Lens, a novel framework to detect concept drift using BERT based embedding feature extraction and output per label model distributions. The work by \cite{feldhans2021drift} studies the performance drift detectors built for low dimensional sensor data on state-of-the-art text classifiers which involves high-dimensional document embeddings. The drift detectors studied are based on Statistical tests such as the Kernel Two-Sample Test, Least-Squares Density Difference and Kolmogorov–Smirnov Test. Through their experiments they highlight the need for drift detectors specifically tailored to text data with high-dimension document embeddings. \cite{zheng2020out} proposes a novel method to generate high quality pseudo Out-Of-Distribution (OOD) samples to effectively improve OOD detection in NLU based tasks. \cite{hendrycks2016baseline} suggests using softmax prediction probabilities to detect OOD samples for text categorization tasks. 
There is a recent toolkit Alibi detect \cite{alibidetect} which proposes solutions for outlier, adversarial and drift detection. However their solution are limited to using pre-trained embeddings to identify drift in text data. \cite{baier2021detecting} introduce the algorithm Uncertainity Drift Detection based on the uncertainty estimates provided by a deep neural network with a Monte Carlo Dropout for classification and regression tasks.

\section{Conclusion and Future Work}
\label{sec:conc}
In this paper, we introduced a tool DetAIL to detect and analyze drift in text. We showed that our approach can compute drift in very early stages and thus can aid in adaptively re-training of models rather than following specific schedules. In addition to this, it helps us in proactively listing out different possibilities of this drift in the form of sample-level and dataset-level explanations which helps in taking necessary action to repair a model. As a future work, we look at how to use these explanations in repairing an NLP Model.

\bibliography{references}

\end{document}